\documentclass{article}

\usepackage[square,sort,comma,numbers]{natbib}
\usepackage[final]{neurips_2022}

\usepackage[utf8]{inputenc} 
\usepackage[T1]{fontenc}    
\usepackage{hyperref}       
\usepackage{url}            
\usepackage{booktabs}       
\usepackage{amsfonts}       
\usepackage{nicefrac}       
\usepackage{microtype}      
\usepackage{times}
\usepackage{latexsym}
\usepackage{graphicx}
\usepackage{enumitem}
\usepackage{array}
\usepackage[dvipsnames]{xcolor}
\usepackage{graphics}
\usepackage{wrapfig}
\usepackage{mathtools}
\title{Topic Segmentation in the Wild: Towards Segmentation of Semi-structured \& Unstructured Chats}

\author{%
  Reshmi Ghosh\textsuperscript{1}\thanks{Corresponding Author; Email: reshmighosh@microsoft.com}\\
  \And
  Harjeet Singh Kajal\textsuperscript{2}\thanks{Work done in collaboration with Microsoft Corporation when a graduate student at UMass, Amherst}\\
  \And
  Sharanya Kamath\textsuperscript{2}\\
  \And
  Dhuri Shrivastava\textsuperscript{2}\\
  \AND
  Samyadeep Basu\textsuperscript{3}\thanks{Work done while at Microsoft Corporation}\
  \AND
  \hspace{-1.5em}
  Soundararajan Srinivasan\textsuperscript{1}\\
  1: Microsoft Corporation; 2: University of Massachusetts, Amherst; 3: University of Maryland
}
\begin{document}

\maketitle
\begin{abstract}
Breaking down a document or a conversation into multiple contiguous segments based on its semantic structure is an important and challenging problem in NLP, which can assist many downstream tasks. However, current works on topic segmentation often focus on segmentation of structured texts. In this paper, we comprehensively analyze the generalization capabilities of state-of-the-art topic segmentation models on unstructured texts. We find that: (a) Current strategies of pre-training on a large corpus of structured text such as Wiki-727K \textit {do not help} in transferability to unstructured texts. (b) Training from scratch with only a relatively small-sized dataset of the target unstructured domain improves the segmentation results by a significant margin.
\end{abstract}

\vspace{-0.6em}

\section{Introduction}
\vspace{-0.8em}
Topic Segmentation refers to the task of splitting texts into meaningful segments that correspond to a distinct topic or a subtopic. Natural language texts, especially in unstructured formats such as chat conversations and transcripts, often do not have an easy-to-detect separation between contiguous topics. Reliable \& accurate division of text into coherent segments can help in making the text more readable as well as searchable. Hence, topic segmentation enables numerous applications such as search assistance and recommendation. It has also been noted that text segmentation can improve and speedup applications such as information extraction and summarization \citep{koshorek2018text}. 

Recently, multiple approaches leveraging language models have been proposed for topic segmentation \citep{koshorek2018text, lukasik2020text, glavavs2021training}. However, the datasets on which these approaches have been evaluated are often structured in nature such as Wiki-727K \citep{koshorek2018text}, Wiki-50  \cite{koshorek2018text}, RST, and Choi \cite{choi2000advances}. Adding to the constraints, in many applications, texts that need to be segmented are often unstructured such as chat transcripts and conversations. But, understanding the effectiveness of topic segmentation methods on such unstructured texts hasn't been well studied. In this paper, we empirically investigate the effectiveness of various topic segmentation methods on unstructured segmentation datasets such as LDC BOLT chat \cite{BOLT} and Amazon Topical chat \cite{gopalakrishnan2019topical}. In addition to being less structured than the Wiki-727K or Wiki-50 data, these datasets are challenging for traditional topic segmentation approaches as they contain grammatically ill-formed "noisy sentences," and a varying number of segments per conversation.  Hence, we systematically examine the effectiveness of large-scale pre-training, dataset used in pre-training, and fine-tuning strategies on these "out-of-domain" (data that is conversational in nature rather than the segmented Wiki content), unstructured text-segmentation datasets spanning traditional the LSTM-based models and the newer transformer-based architectures. 

We find that large-scale pre-training (and fine-tuning with data from target domain) has only a {\it negligible} effect on downstream segmentation tasks, when the task consists of unstructured data. This is contrary to the conventional wisdom in NLP, where pre-training and fine-tuning is a common practice. We, therefore, identify topic segmentation on unstructured data as one task where large-scale pre-training doesn't have any significant effect. To perform well on segmentation of unstructured text, we find that training, from scratch, with only a {\it few} examples of the segmentation domain is sufficient. This is true for segmentation architectures ranging from the LSTM-based ones to the recent Transformer-based ones.  Our results also show that, for unstructured topic segmentation, avoiding pre-training on a large corpus such as the Wiki-727K dataset results in saving a significant amount of training time and compute resources, facilitating the exploration of newer approaches. In summary, our contributions are as follows:

\begin{itemize}[noitemsep,topsep=0pt,parsep=0pt,partopsep=0pt]
    \item We investigate and stress-test the effectiveness of current topic segmentation methods on unstructured texts, which is a more challenging segmentation task when compared to structured-segmentation tasks. 
    \item We find that pre-training on large topic segmentation datasets such as Wiki-727K has negligible effect on downstream transfer to unstructured text-segmentation datasets and instantiating the model with only a few-examples of the unstructured task is sufficient. 
    \item We provide simple and practical recipes to improve performance on unstructured-segmentation datasets. 
\end{itemize}
\vspace{-0.6em}
\section{Related Works}\label{related_works}
\vspace{-0.8em}
Topic segmentation has been explored through many realms; particularly, the approaches used could be broadly categorized as Non-neural-based and Neural-based approaches in both supervised and unsupervised settings.

\textbf{Non-neural approaches}:
The early research efforts related to non-neural approaches for topic segmentation include \cite{TextTiling} that focused on an unsupervised approach to analyze lexical cohesion in small segments by leveraging counts of word repetitions. This work was expanded to enable models understand words and sentences occurring in segments in a comprehensive manner leading to the wide use of lexical chains \cite{adarve2007topic, sitbon2007topic}.

\textbf{Neural Approaches:}\label{non_neural} \cite{koshorek2018text} used a hierarchical Bi-LSTM to cast the topic segmentation as a supervised learning task. Other neural methods also leverage the Transformer architecture. In \cite{lukasik2020text}, the authors proposed three transformer-based models, of which Cross-Segment BERT model is particularly important for topic segmentation tasks as the model captures information from the local context surrounding a potential topic segment boundary to judge about which pool of sub-document units, the particular segment belongs. The other two model architectures use a hierarchical approach as used by \cite{koshorek2018text}, but with using the BERT model instead of BiLSTMs. \cite{DBLP:journals/corr/abs-2106-12978} is another recent work that uses unsupervised approach based on BERT embeddings to segment topics in multi-person meeting transcripts.


\vspace{-0.8em}
\section{Segmentation Datasets}
\vspace{-0.8em}
\textbf{Unstructured Datasets}: In this work, We use the LDC's (Linguistic Data Consortium) BOLT SMS and chat data collection (filtered to only include chats and SMS in English; hereafter called as BOLT) and Amazon Topical Chat dataset. The BOLT dataset is significantly more unstructured in nature than Topical chat dataset, as the conversations in BOLT chat contain non-uniform sentence structure, incomplete sentences, and abbreviations. Thus, in this paper we refer to the BOLT data as unstructured dataset, and label the Topical chat dataset as semi-structured dataset (as it has more well-formed sentence structure; additional details and snapshots of the datasets in Appendix \ref{dataset}).

\textbf{Wiki-727K}: Proposed by \cite{koshorek2018text}, this dataset contains 727,746 documents in the English language with text segmentations created according to the table of contents. The Wiki-727K dataset is considered a structured dataset because the text in it is not conversational in nature and has proper syntactical structure in the form of well-formed sentences grouped into paragraphs and sections.



These datasets were pre-processed (Appendix sub-section \ref{appdx:preprocess}) to generate labels for each sentence in every segment, for use in the supervised learning formulation.

\vspace{-0.6em}
\section{Experiments}\label{experiments}
\vspace{-0.8em}
We study the problem of segmenting semi-structured and unstructured chats using three popular modeling paradigmns used in the structured segmentation domain: the Hierarchical Bi-LSTM model proposed by \cite{koshorek2018text} and the Cross-Segment BERT model \cite{lukasik2020text} (hereafter CSBERT), and the Cross-Segment RoBERTa (CSRoBERTa)  \cite{glavavs2021training} (Appendix \ref{sec:appendix}). In CSRoBERTa, we use the same training paradigm as CSBERT, but replace BERT with RoBERTa-base. 
Note that these models in the existing body of work have not been validated against any form of semi-structured or unstructured datasets. Table \ref{tab:exp4} presents the results of our analysis. We first pre-train all three models on Wiki-727K and test against the unstructured BOLT, the semi-structured Topical Chat, and the structured Wiki-727K datasets (underlined in Table \ref{tab:exp4}). We then use different pre-training and fine-tuning combinations to examine the necessity of large-scale pre-training on structured datasets for segmenting conversational data. Also note that the results in Table \ref{tab:exp4} are generated by partitioning the BOLT and Topical Chat data into documents of 5 segments each (validated in Figure \ref{fig1}; more details on train-test split are in Appendix \ref{appdx:exp_setup}).

\vspace{-0.4em}
Adapting the hierarchical Bi-LSTM model on the unstructured BOLT and the semi-structured Topical chat datasets during the validation phase, we conclude that the F1 scores (additional details on evaluation metrics in Appendix subsection \ref{appdx:metrics}) are significantly worse when compared with the performance of the CSBERT and CSRoBERTa model in the same setting. However, evaluating the performance on the structured Wiki-727K dataset, we find that the F1 scores from both models are in the same range. 
\vspace{-0.8em}
\begin{center}
\begin{table*}

\scalebox{0.8}{
\hspace{-0.9cm}
\begin{tabular} {  c|c|c|c|c|c|c}
\hline
&

\multicolumn{3}{c|}{\small \textbf{Datasets}}

&
&
&
\\
\hline
\footnotesize \textbf{Task} 
&
\footnotesize \textbf{Pre-Train} 
&
\footnotesize \textbf{Finetune}
&
\footnotesize \textbf{Test}
&
\footnotesize \textbf{Cross Segment BERT}
&
\footnotesize \textbf{Cross Segment RoBERTa-Base}
&
\footnotesize \textbf{Hierarchical Bi-LSTM}
\\
\hline
\footnotesize A.1 & \footnotesize Wiki-727K &  \footnotesize - & \footnotesize Topical Chat & \footnotesize 0.492 & \footnotesize 0.487 &  \footnotesize 0.021\\
\footnotesize A.2 &\footnotesize Wiki-727K & \footnotesize BOLT & \footnotesize Topical Chat & \footnotesize 0.470 & \footnotesize 0.406 & \footnotesize 0.391 \\
\footnotesize A.3 & \footnotesize Wiki-727K & \footnotesize Topical Chat & \footnotesize Topical Chat & \footnotesize \underline{0.725} & \footnotesize \underline{0.713} & \footnotesize \underline{0.931} \\
\hline
\footnotesize A.4 & \footnotesize BOLT & \footnotesize - & \footnotesize Topical Chat & \footnotesize 0.491 & \footnotesize  0.498 & \footnotesize 0.611\\
\footnotesize A.5 & \footnotesize BOLT & \footnotesize Topical Chat & \footnotesize Topical Chat & \footnotesize 0.734 & \footnotesize 0.729 & \footnotesize 0.915\\
\hline
\footnotesize A.6 & \footnotesize Topical Chat & \footnotesize - &  \footnotesize Topical Chat &  \footnotesize 0.764 &   \footnotesize  \bf\textcolor{purple}{0.767} &  \footnotesize  \bf\textcolor{purple}{0.951}\\
\footnotesize A.7 & \footnotesize Topical Chat & \footnotesize BOLT & \footnotesize Topical Chat & \footnotesize  \bf\textcolor{purple}{0.767} & \footnotesize 0.759 & \footnotesize 0.501\\
\hline
\hline

\footnotesize B.1 & \footnotesize Wiki-727K &  \footnotesize - & \footnotesize BOLT & \footnotesize 0.487 & \footnotesize 0.467 &  \footnotesize 0.005\\
\footnotesize B.2 & \footnotesize Wiki-727K & \footnotesize BOLT & \footnotesize BOLT & \footnotesize \underline{0.489} & \footnotesize \footnotesize \underline{0.479} & \footnotesize \underline{0.406} \\
\footnotesize B.3 & \footnotesize Wiki-727K & \footnotesize Topical Chat & \footnotesize BOLT & \footnotesize 0.511 & \footnotesize 0.492 & \footnotesize 0.152 \\
\hline
\footnotesize B.4 & \footnotesize BOLT & \footnotesize - & \footnotesize BOLT & \footnotesize \bf\textcolor{cyan}{0.569} & \footnotesize \bf\textcolor{cyan}{0.561} & \footnotesize  \bf\textcolor{cyan}{0.443}\\
\footnotesize B.5 & \footnotesize BOLT & \footnotesize Topical Chat & \footnotesize BOLT & \footnotesize 0.518 & \footnotesize 0.509 & \footnotesize 0.181\\
\hline
\footnotesize B.6 & \footnotesize Topical Chat & \footnotesize - &  \footnotesize BOLT &  \footnotesize 0.544 &  \footnotesize  0.542 &  \footnotesize 0.157\\
\footnotesize B.7 & \footnotesize Topical Chat & \footnotesize BOLT & \footnotesize BOLT & \footnotesize 0.536 & \footnotesize 0.529 & \footnotesize 0.331\\

\hline

\hline

\hline
\footnotesize C.1 & \footnotesize Wiki-727K &  \footnotesize - & \footnotesize Wiki-727K & \footnotesize {\bf\textcolor{orange}{0.604}} & \footnotesize {\bf\textcolor{orange}{0.599}} &  \footnotesize {\bf\textcolor{orange}{0.57}}\\
\footnotesize C.2 & \footnotesize Wiki-727K & \footnotesize BOLT & \footnotesize Wiki-272K & \footnotesize 0.433 & \footnotesize 0.435 & \footnotesize 0.501 \\
\footnotesize C.3 & \footnotesize Wiki-727K & \footnotesize Topical Chat & \footnotesize Wiki-727K & \footnotesize 0.513 & \footnotesize 0.509 & \footnotesize 0.411 \\
\hline

\footnotesize C.4 & \footnotesize BOLT & \footnotesize - & \footnotesize Wiki-727K & \footnotesize 0.487 & \footnotesize  0.492 & \footnotesize 0.12 \\
\footnotesize C.5 & \footnotesize BOLT & \footnotesize Topical Chat & \footnotesize Wiki-727K & \footnotesize 0.489 & \footnotesize 0.478 & \footnotesize 0.027 \\
\hline

\footnotesize C.6 & \footnotesize Topical Chat & \footnotesize - &  \footnotesize Wiki-727K &  \footnotesize 0.505 &  \footnotesize 0.502 &  \footnotesize 0.020\\
\footnotesize C.7 & \footnotesize Topical Chat & \footnotesize BOLT & \footnotesize Wiki-727K & \footnotesize 0.5089 & \footnotesize 0.511 & \footnotesize 0.198\\
\hline
\end{tabular}}


\vspace{-0.5em}
\caption{\footnotesize Effect of model architectures and training strategies on topic segmentation tasks, grouped by the test dataset of choice - Topical Chat {\bf(A.1 - 1.7)}, BOLT  {\bf(B.1 - B.7)}  \& Wiki-727K {\bf(C.1 - C.7)}. The best F1 scores for \textcolor{cyan}{\bf{BOLT}} (unstructured), \textcolor{purple}{\bf{Topical Chat}} (semi-structured), and \textcolor{orange}{\bf{Wiki-727K}}, per model are highlighted. Comparing the F1 scores in different pre-training \& fine-tuning scenarios, we conclude that the Cross-Segment BERT model {\it consistently} outperforms the Hierarchical Bi-LSTM, and CSRoBERTa model on {\it almost} all tasks. We also find that large-scale pre-training with the structured Wiki-727K dataset (and then fine-tuning with data from Target domain - underlined; Task B.2 vs. B.4 and Task A.3 vs. A.6) is not required to create cohesive topic segments on unstructured or semi-structured conversational data.(Train-finetuning-test split described in Appendix \ref{appdx:exp_setup}).}
\label{tab:exp4}

\end{table*}
\end{center}
\vspace{-0.5em}

\subsection{Effectiveness of Pre-training on Wiki-727K}
\vspace{-0.8em}
In this section, we investigate the effectiveness of pre-training on the large Wiki-727K dataset. We first pre-train the Hierarchical Bi-LSTM, CSRoBERTa, and CSBERT models on the 80\% of the Wiki-727K corpus. We further fine-tune the models on the unstructured datasets: BOLT chat and Topical Chat dataset. Additionally, we experiment with fine-tuning the models after training on semi-structured/unstructured dataset instead of the pre-training step with Wiki-727K. From Table \ref{tab:exp4}, we find that training models with unstructured/semi-structured and fine-tuning it with semi-structured and unstructured dataset respectively, leads to better performance than fine-tuning with the Wiki-727K checkpoint. For instance, we find that with the pre-train with Wiki-727K and fine-tune paradigm, using the CSBERT, we obtain a F1 score of 0.725 for Topical Chat, whereas only training on Topical Chat results in a F1 score of 0.764 ({\it Task A.3 vs.A.6})\footnote{We find that for Cross-Segment BERT model Task A.6 and A.7 have very similar F1 scores}. For BOLT, the pre-train and fine-tune paradigm results in a F1 score of 0.489 whereas training from scratch with BOLT dataset results in a F1 score of 0.499 ({\it Task B.2 vs. B.4}). These results indicate that the conventional approach of pre-training on a large structured Wiki-727K dataset, and then fine-tuning with semi-structured or unstructured dataset, doesn't lead to an improvement in F1 scores, making the approach of pre-training on structured dataset questionable. We associate this finding to the fact that - Wiki-727K, although large enough for pre-training approaches and has been used in established Topic Segmentation methods for structured texts, the dataset fails to represent the rapid change in themes of conversations, thus making feature reuse \citep{neyshabur2020being} from the pre-training process redundant. In chats between two human agents, the topic of the conversation can change very quickly, which is not fully represented by feature hierarchy learned from structured texts.

\vspace{-0.9em}
\subsection{Effect of architecture on unstructured datasets}
\vspace{-0.8em}
We test our initial conclusions (Table \ref{tab:exp4} on the efficacy of architectures across various number of segments. We find that, across different number of segments, the CSBERT model performs substantially better than the Hierarchical Bi-LSTM model and slightly better than CSRoBERTa model. From Figure  \ref{fig1}, we conclude that:

\begin{wrapfigure}{l}{0.5\textwidth}
\hspace{-1.5em}
    \centering
 	\includegraphics[width = 0.49\textwidth]{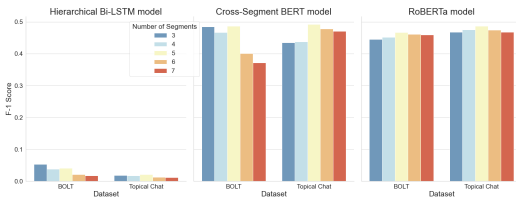}
 	\caption{ \footnotesize 
 	Effect of model architecture and the number of segments on segmentation performance. Comparing the F1 scores from all three models pre-trained on Wiki-727K and inferred on unstructured \& semi-structured datasets (without any fine-tuning), we find that CSBERT outperforms the other models robustly across the varying number of segments. Bi-LSTM performs very poorly compared to the other two models, which can be explained by its inability to fully consider the semantic context of the text representing segment boundaries. Additionally, with an increase in the number of segments, the F1 score peaks at 5 segments, and then drops across all models. We attribute this finding to the fact that at 5 segments, we are able to capture a relatively decent break in the theme of conversations for both datasets.} 
 	\label{fig1}
\end{wrapfigure}

\vspace{-0.8em}

\begin{itemize}[noitemsep,topsep=0pt,parsep=0pt,partopsep=0pt]
\item Across the varying number of segments curated and adapted during the pre-training step with Wiki-727K, the CSBERT model results in higher F1 scores when tested against semi-structured Topical Chat and unstructured BOLT datasets. Hence, we can conclude that for these topic segmentation tasks, CSBERT model is more suitable than the Hierarchical Bi-LSTM and the CSRoBERTa models.
\item As the number of segments in the conversational datasets increases, the F1 scores from all models drop. We associate this finding with the fact that the large segments of the conversational data may contain more heterogeneous topics, making it difficult for all models to group coherent chats.
\end{itemize}



\vspace{-0.8em}
\subsection{Practical recipes to improve unstructured segmentation tasks}
\vspace{-0.8em}
Casting topic segmentation for unstructured datasets as a binary classification problem leads to a severe imbalance in class labels, prompting the need to re-weight the samples or even modify the loss function to boost performance. The number of end-of-segment sentences (encoded as `1') is significantly smaller than the non-boundary sentences (sentences that do not mark the end of a segment; encoded as label `0') due to the inherent structure of segments in any document or chat. 
\vspace{-0.7em}
\subsubsection{Re-weighting in cross-entropy loss}
\vspace{-0.8em}
To reduce the effect of dominance by the samples with label `0' and avoid biasing the model at inference time, we re-weight the class labels in cross-entropy loss function, giving proportional importance to labels `0' and `1'. The set of weights to be assigned is considered as a hyper-parameter and is optimized in the range {$[0, 1]$}.


We find that weighting the end-of-segment sentences (encoded as `1') with 0.8, and weighting the rest with 0.2 yields the best results on these datasets. From Table \ref{tab:exp9}, we conclude that re-weighting the cross-entropy loss function to provide proportional importance to both labels leads to a slightly better F1 score for all three models.

\vspace{-0.7em}
\subsubsection{Focal loss as an alternative loss function for imbalanced topic segmentation}\label{Focallossdescp}
\vspace{-0.8em}
Focal loss has been used widely to mitigate the risks involved with class imbalance for tasks related to object detection \cite{lin2017focal}, credit-card fraud detection \cite{focalloss}, and other tasks involving class-imbalance \cite{mukhoti2020calibrating}. We consider the \(\alpha\) (a parameter that controls trade-off between precision and recall) and \(\gamma\) (focusing parameter; defines the degree of confidence assigned by the model to correct predictions that contributes to overall loss  values) as focal-loss hyper-parameters and tune these over 10 epochs (additional details and equations in Appendix \ref{appdx:practical_recipes}). 



\vspace{-0.8em}
\begin{center}

\begin{table*}
\scalebox{0.9}{
\hskip -0.4cm
\begin{tabular} {c|c|c|c|c|c|c|c|c}
\hline
\multicolumn{3}{c|}{\small \textbf{Datasets}}
&
\multicolumn{3}{c|}{\small \textbf{Cross-Entropy Weights = $[0.2, 0.8]$}}

&
\multicolumn{3}{c}{\small \textbf{Focal Loss (\small {\(\alpha\) = 0.8 ; \(\gamma\) = 2)}}}
\\
\hline
\footnotesize \textbf{Pre-Train} 
&
\footnotesize \textbf{Finetune}
&
\footnotesize \textbf{Test}
&
\footnotesize \textbf{CSBERT}
&
\footnotesize \textbf{CSRoBERTa}
&
\footnotesize \textbf{Bi-LSTM}
&
\footnotesize \textbf{CSBERT}
&
\footnotesize \textbf{CSRoBERTa}
&
\footnotesize \textbf{Bi-LSTM}
\\
\hline
\footnotesize Wiki-727K & \footnotesize BOLT & \footnotesize BOLT & \footnotesize 0.495 & \footnotesize 0.481 & \footnotesize 0.409 &  \footnotesize 0.503 &  \footnotesize  0.485 &  \footnotesize 0.432\\
\footnotesize BOLT &  \footnotesize - & \footnotesize BOLT & \footnotesize 0.575 & \footnotesize 0.567 &  \footnotesize 0.443 &  \footnotesize 0.580 &  \footnotesize 0.572 &  \footnotesize 0.45\\
\hline
\footnotesize Wiki-727K & \footnotesize TC & \footnotesize TC & \footnotesize 0.729 & \footnotesize 0.717 & \footnotesize 0.933 &  \footnotesize 0.748 &  \footnotesize 0.729 &   \footnotesize 0.928\\
\footnotesize TC &  \footnotesize - & \footnotesize TC & \footnotesize 0.767 & \footnotesize 0.769 &  \footnotesize 0.952 &  \footnotesize 0.778 &  \footnotesize 0.775&  \footnotesize 0.95\\
\hline
\end{tabular}
}

\vspace{-0.5em}
\caption{\footnotesize Results from models  trained with re-weighted cross-entropy loss and focal-Loss function.CSBERT refers to results obtained using the cross-segment BERT model.Additional results from all training strategies, as illustrated in Table \ref{tab:exp4} can be found in Appendix \ref{appdx:practical_recipes}, Table \ref{tab:exp10}.}
\label{tab:exp9}
\end{table*}

\end{center}
\vspace{-0.8em}
From Table \ref{tab:exp9}, we find that the resultant F1 scores with focal loss are higher than re-scaling the cross-entropy loss function. Hence, we conclude that for future iterations of topic segmentation tasks involving unstructured datasets, focal loss is a better alternative.
\vspace{-0.8em}
\section{Conclusion}
\vspace{-0.8em}
In this work, we analyzed the effectiveness of current topic segmentation methods on unstructured texts. We find that, across different architectures and datasets, pre-training on the large, structured Wiki-727K {\it is not required} for segmentation of unstructured datasets such as BOLT and Topical Chat that contain syntactical and semantical noise. Training from scratch with only a few-examples provides a sufficiently strong baseline for this task. This finding is counter to the pre-training on a large corpus and fine-tuning on the target domain paradigm, commonly used in a variety of tasks in NLP. Further, we provided various practical fine-tuning recipes to boost the segmentation performance on such conversational and unstructured datasets.

{
\bibliography{main}
\bibliographystyle{plain}

\appendix

\section{Appendix}

\subsection{Dataset}
\label{dataset}

To the best of our knowledge, Wiki-727K is the only publicly available large dataset, whose size makes it appropriate for large-scale pretraining. In fact, no singular conversational (unstructured) data of that scale exists, that can be used to pre-train large Transformer model. A similar approach of pre-training was undertaken in \cite{koshorek2018text, lukasik2020text, glavavs2021training}. \\

\begin{figure}[!ht]
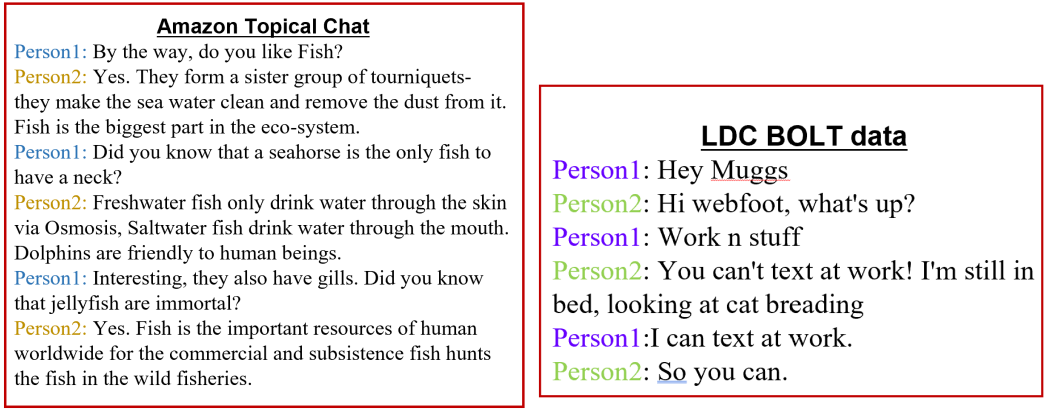

    \includegraphics[width=.5\textwidth]{Images/Topical_Chat_dataset.pdf}\hfill
    \includegraphics[width=.5\textwidth]{Images/LDC_BOLT.pdf}\hfill

    \caption{Snapshot of Amazon Topical Chat and LDC BOLT datasets. The former dataset contains conversations which have well-formed sentences, whereas the latter contains conversations which have a higher degree of 'unstructured-ness', i.e., incomplete sentences and abbreviations. The LDC BOLT dataset more closely represents the modern-day fast paced SMS/text conversations. }\label{fig:appdx;}
\end{figure}

BOLT SMS and Chat dataset by LDC have conversations that have been extracted from messaging platforms like WhatsApp, iMessage, Android SMS, Symbian SMS, Viber, BlackBerry, QQ, Google chat, Skype chat, \& Yahoo Messenger in languages like Chinese, Egyptian Arabic and English. The dataset contains 2140 Egyptian, 7844 Chinese and 9155 English conversations. We work with only the English portion of the data. Additionally, the dataset have the following characteristics:
         \begin{itemize}
         \item Mean sentence length (number of words) = 9.45 ; standard deviation = 9.41
         \item Mean segment length (number of sentences) = 11.28; standard deviation  = 13.99
         \end{itemize}
         
Topical Chat dataset from Amazon contains human to human conversations in 8 broad topics. It has more than 8000 conversations. Even though this dataset has well-formed sentence structure, it is in the form of conversations, and not articles, and thus we categorize the dataset as semi-structured. The underlying knowledge in Topical Chat spans across 8 broad topics and there is very little to no variation in the segment length of this dataset. 
         \begin{itemize}
         \item Mean sentence length (number of words) = 19.86, standard deviation  = 10.49
         \item Mean segment length (number of sentences) = 21.83, standard deviation  = 1.75
        \end{itemize}

\subsection{Pre-Processing}\label{appdx:preprocess}
Both Topical chat and BOLT datasets are conversational in nature, and thus required additional pre-processing. We achieve this by segmenting these datasets into multiple segments (we used 5 segments, as it resulted in best F1 scores; {\it Figure \ref{fig1}}), wherein each segment is a snippet of a specific chat conversation. Furthermore, we cast the problem of topic segmentation for OOD datasets as a supervised learning problem, and thus we create synthetic labels for each sentence in every segment. We first pre-process the datasets to synthetically create multiple segments, and label the sentences occurring in each segment $x_i = {(x_1, x_2, ...., x_n)}$ according to the boundary condition, i.e., whether a specific sentence was end-of-segment (label, $y_i =$ `1') or non end-of-segment sentence (encoded label $y_i =$ `0').

\subsection{Experimental Setup}
\label{appdx:exp_setup}

For the initial task of pretraining on large structured dataset, we first randomly partitioned the Wiki-727K dataset in 80\% / 10\% / 10\% format to create train, development (fine-tuning), and test set. We then used train set from Wiki-727K dataset, to perform the pretraining using Hierarchical Bi-LSTM, Cross-Segment BERT, and Cross-Segment RoBERTa models.\\ 

Additionally for the initial task of pre-training on Wiki-727K and testing on semi-structured Topical Chat, and unstructured LDC dataset, we first partition all conversations to form segments (snippets of conversations) and group the segment chunks to form documents. The model at any point of the training process consumed a batch of these documents. As described in section \ref{appdx:preprocess}, the sentences in these segments are synthetically labelled to  indicate the end-of-segment. As a result, the chunking the conversations from BOLT and Topical Chat datasets in 5 segments leads to 1815 and 1726 documents, respectively. \\

Furthermore, the documents created from the chunking process described above were split into train/fine-tuning/test sets for tasks described in Table \ref{tab:exp4} and Table \ref{tab:exp10}. The documents from Topical Chat dataset was divided in 1099 / 348 / 281 documents for the train/fine-tuning/test splits respectively, wherein each document had 5 segments of mean segment length of approximately 11.5 sentences. Additionally, a similar approach was employed for the documents from BOLT dataset, and it was divided in 1090 / 363 / 362 documents for the train/fine-tuning/test splits respectively containing 5 segments each (mean segment length of ~21.83 sentences). \\


Lastly to train the three models from scratch using the unstructured, and semi-structured data, without involving any pre-training on structured Wiki-727K, required 1109 documents of 5 segments from BOLT, and 1030 documents of 5 segments from Topical Chat. 

\subsection{Model Architecture}
\label{sec:appendix}
As detailed in Section \ref{related_works}, topic segmentation in the existing body of work has involved neural and non-neural approaches, but owing to the complexity of understanding heterogeneous conversational dataset, we use state-of-the-art neural models, i.e, Hierarchical Bi-LSTM as proposed by \cite{koshorek2018text}, and CSBERT model, introduced by \cite{lukasik2020text}.\\

The hierarchical Bi-LSTM model first learns the sentence representation and is then fed into a segment prediction sub-network. The lower-level sub-network has two-layer bi-directional LSTM layer that generates sentence representations by taking in words $w\textsubscript{1}, w\textsubscript{2}....w\textsubscript{i}$ of a sentence $x\textsubscript{i}$ as input to the model, and the intermediate output is passed through a max pooling layer to create the sentence representations $e\textsubscript{i}$. The upper-level sub-network for segment prediction then takes a sequence of sentence embeddings generated from the previous layers, and feeds them to a two-layer Bi-LSTM layer, which then feeds into a fully connected-layer with softmax function to generate segmentation probabilities.\\

In the Cross Segment BERT model, the authors leverage information from the local context, i.e., studying the semantic shift in word distributions, as first introduced in \cite{TextTiling}. The additional context on both sides of the segment boundary (termed as `candidate break in the paper), i.e., the sequence of word-piece tokens that come before and after the segment breaks. Basically the model is fed $k$ word-piece tokens from the left and $k$ tokens from the right of a segment break. The input is composed of a classification token (denoted by $[CLS]$ ), followed by the two contexts concatenated together, and separated by a separator token (denoted by $[SEP]$ ). The tokens are fed into the Transformer encoder(\cite{Vaswani}) which is initialized by $BERT\textsubscript{LARGE}$ to output segmentation probabilities.  The $BERT_\textsubscript{LARGE}$ model has 24 layers and uses 1024-dimensional embeddings and 16-attention heads. As the released BERT checkpoint only supports up to 512 tokens, we keep at maximum 250 word-tokens on each side.

The RoBERTa architecture was chosen as a comparative alternative to the BERT model in the Cross-Segment framework, as it is relatively newer successor of BERT \cite{liu2019roberta}. However, rather than changing the framework of Cross-Segment learning as proposed by \cite{lukasik2020text}, wherein the authors demonstrate the capability of BERT model to learn the context around end-of-segments in a robust way, we chose to utilize the same framework and simply replace the BERT model by RoBERTa-base.

\subsection{Evaluation Metrics}\label{appdx:metrics}

Since we cast the task of topic segmentation as a binary classification problem, we use Precision, Recall and F1 scores to measure performance. Precision measures the percentage of boundaries identified by the model that are true boundaries. Recall measures the percentage of true boundaries identified by the model. Although comprehensive, it is important to note that there are some challenges associated with individually reporting Precision and Recall as they are somewhat less sensitive to near misses of true boundary identifications, when the prediction is off by one or two sentences, and thus we report F1 scores in Section \ref{experiments}. F1 score can be reliably used to conclude our initial findings from performing topic segmentation based binary classification on unstructured and semi-structured data.

\subsection{Practical Recipes: Additional Results}\label{appdx:practical_recipes}
Focal loss(\ref{second_eqn}) is different from Cross Entropy loss ( \ref{first_eqn}), as the former implements a technique called as "down-weighting", that reduces the influence of confidently predicting easy examples (predicted probability: $p >> 0.5$) on the loss function, resulting in more attention being paid to hard-to-predict examples (misclassified examples). To achieve this, additional modulating-factor, called the focusing parameter ($\gamma$) is included to improve the conventional Cross Entropy loss function. Additionally, Focal loss also tackles the class-imbalance problem by introducing a weighting parameter ($\alpha$) to place appropriate weights on positive and negative classes.

\begin{equation}
    CE(p,y) =
\begin{cases}
    -log(p),& \text{if } y = 1\\
    -log(1-p),              & \text{otherwise}
\end{cases}
\label{first_eqn}
\end{equation}

\begin{equation}
    FL(p,y) = 
    \begin{cases}
    -\alpha(1-p)^{\gamma}log(p), & \text{if } y = 1 \\
    -(1-\alpha)p^{\gamma}log(1-p), & \text{otherwise}
    \end{cases}
\label{second_eqn}
\end{equation}

Thus, the Focal Loss function is a dynamically scaled Cross Entropy loss, where the scaling factor decays to zero as confidence in the correct class increases.\\

In Table \ref{tab:exp10}, we present our findings by experimenting with re-weighted cross-entropy loss and Focal loss functions for all combinations of pre-training and fine-tuning strategies. We find that, across {\it almost} (except 3) all paradigms, Cross-Segment BERT model outperforms Cross-Segment RoBERTa and Hierarchical Bi-LSTM model on both re-weighting and focal loss function recipes. Hierarchical Bi-LSTM F1 scores are largely in-consistent, and thus we can conclude the re-weighting of cross-entropy loss and replacement by focal loss function in our baseline model did not prove of large significance to the Bi-LSTM framework. We can attribute this finding to the RNN structure which limits contextual learning around the segment boundaries, whereas Transformer-based architectures, with their multi-head attention mechanism learn about the context robustly.\\

Moreover, Focal loss across all pre-training and fine-tuning strategies prove to be a robust alternative than re-weighting cross-entropy loss, for all three models. We hypothesize this effect to be a result of including appropriate values of focusing parameter($\gamma$) and trade-off parameter($\alpha$) in the Focal loss function, which assigns la

\begin{center}
\begin{table*}[htp]
\scalebox{0.9}{
\hskip -1cm
\begin{tabular} {c|c|c|c|c|c|c|c|c}
\hline
\multicolumn{3}{c|}{\small \textbf{Datasets}}
&
\multicolumn{3}{c|}{\small \textbf{CE weights = [0.2, 0.8]}}

&
\multicolumn{3}{c}{\small \textbf{Focal Loss (\small \(\alpha\) = 0.8; \(\gamma\) = 2)}}
\\
\hline
\footnotesize \textbf{Pre-Train} 
&
\footnotesize \textbf{Finetune}
&
\footnotesize \textbf{Test}
&
\footnotesize \textbf{CSBERT}
&
\footnotesize \textbf{CSRoBERTa}
&
\footnotesize \textbf{Bi-LSTM}
&
\footnotesize \textbf{CSBERT}
&
\footnotesize \textbf{CSRoBERTa}
&
\footnotesize \textbf{Bi-LSTM}
\\
\hline
\footnotesize Wiki-727K &  \footnotesize - & \footnotesize Topical Chat & \footnotesize 0.497 & \footnotesize 0.491 &  \footnotesize 0.028 &  \footnotesize 0.512 &  \footnotesize 0.504&  \footnotesize 0.037\\
\footnotesize Wiki-727K & \footnotesize BOLT & \footnotesize Topical Chat & \footnotesize 0.475 & \footnotesize 0.411 & \footnotesize 0.398 &  \footnotesize 0.483 &  \footnotesize 0.427 &  \footnotesize 0.405\\
\footnotesize Wiki-727K & \footnotesize Topical Chat & \footnotesize Topical Chat & \footnotesize 0.729 & \footnotesize 0.717 & \footnotesize 0.933 &  \footnotesize 0.748 &  \footnotesize 0.729 &  \footnotesize 0.928  \\
\hline
\footnotesize BOLT & \footnotesize - & \footnotesize Topical Chat & \footnotesize 0.493 & \footnotesize  0.5 & \footnotesize 0.613 &  \footnotesize 0.501&  \footnotesize 0.510&  \footnotesize 0.606\\
\footnotesize BOLT & \footnotesize Topical Chat & \footnotesize Topical Chat & \footnotesize 0.736 & \footnotesize 0.731 & \footnotesize 0.917 &  \footnotesize 0.747&  \footnotesize 0.741&  \footnotesize 0.920\\
\hline
\footnotesize Topical Chat & \footnotesize - &  \footnotesize Topical Chat &  \footnotesize 0.767 &  \footnotesize {\bf\textcolor{purple}{0.769}} &  \footnotesize {\bf\textcolor{purple}{0.952}} &  \footnotesize {\bf\textcolor{purple}{0.778}}&  \footnotesize {\bf\textcolor{purple}{0.775}}&  \footnotesize {\bf\textcolor{purple}{0.95}}\\
\footnotesize Topical Chat & \footnotesize BOLT & \footnotesize Topical Chat & \footnotesize {\bf\textcolor{purple}{0.768}} & \footnotesize 0.761 & \footnotesize 0.505 &  \footnotesize 0.777&  \footnotesize 0.768&  \footnotesize 0.515\\
\hline
\hline
\footnotesize Wiki-727K &  \footnotesize - & \footnotesize BOLT & \footnotesize 0.490 & \footnotesize 0.468 &  \footnotesize 0.007 &  \footnotesize 0.493 &  \footnotesize 0.472 &  \footnotesize 0.012\\
\footnotesize Wiki-727K & \footnotesize BOLT & \footnotesize BOLT & \footnotesize 0.495 & \footnotesize 0.481 & \footnotesize 0.409 &  \footnotesize 0.503 &  \footnotesize 0.485 &   \footnotesize 0.432\\
\footnotesize Wiki-727K & \footnotesize Topical Chat & \footnotesize BOLT & \footnotesize 0.573 & \footnotesize 0.495 & \footnotesize 0.183 &  \footnotesize 0.560 &  \footnotesize  0.524 &  \footnotesize 0.214\\
\hline

\footnotesize BOLT & \footnotesize - & \footnotesize BOLT & \footnotesize {\bf\textcolor{cyan}{0.575}} & \footnotesize {\bf\textcolor{cyan}{0.567}} & \footnotesize {\bf\textcolor{cyan}{0.443}} &  \footnotesize {\bf\textcolor{cyan}{0.580}} &  \footnotesize {\bf\textcolor{cyan}{0.572}}&  \footnotesize {\bf\textcolor{cyan}{0.45}}\\
\footnotesize BOLT & \footnotesize Topical Chat & \footnotesize BOLT & \footnotesize 0.520 & \footnotesize 0.511 & \footnotesize 0.183 &  \footnotesize 0.531 &  \footnotesize 0.519&  \footnotesize 0.189\\
\hline
\footnotesize Topical Chat & \footnotesize - &  \footnotesize BOLT &  \footnotesize 0.546 &  \footnotesize 0.544 &  \footnotesize 0.159 &  \footnotesize 0.555&  \footnotesize 0.551&  \footnotesize 0.164\\
\footnotesize Topical Chat & \footnotesize BOLT & \footnotesize BOLT & \footnotesize 0.537 & \footnotesize 0.531 & \footnotesize 0.333 &  \footnotesize 0.549&  \footnotesize 0.54&  \footnotesize 0.34\\

\hline

\hline

\footnotesize Wiki-727K &  \footnotesize - & \footnotesize Wiki-727K & \footnotesize {\bf\textcolor{orange}{0.609}} & \footnotesize {\bf\textcolor{orange}{0.602}} &  \footnotesize {\bf\textcolor{orange}{0.591}} &  \footnotesize {\bf\textcolor{orange}{0.614}} &  \footnotesize {\bf\textcolor{orange}{0.611}} &  \footnotesize {\bf\textcolor{orange}{0.6}}\\
\footnotesize Wiki-727K & \footnotesize BOLT & \footnotesize Wiki-272K & \footnotesize 0.435 & \footnotesize 0.438 & \footnotesize 0.508 &  \footnotesize 0.441 &  \footnotesize 0.446&  \footnotesize 0.513 \\
\footnotesize Wiki-727K & \footnotesize Topical Chat & \footnotesize Wiki-727K & \footnotesize 0.516 & \footnotesize 0.510 & \footnotesize 0.414 &  \footnotesize 0.522 &  \footnotesize 0.517 &  \footnotesize 0.417\\
\hline
\footnotesize BOLT & \footnotesize - & \footnotesize Wiki-727K & \footnotesize 0.489 & \footnotesize  0.495 & \footnotesize 0.127 &  \footnotesize 0.493&  \footnotesize 0.501&  \footnotesize 0.132\\
\footnotesize BOLT & \footnotesize Topical Chat & \footnotesize Wiki-727K & \footnotesize 0.492 & \footnotesize 0.479 & \footnotesize 0.03 &  \footnotesize 0.496&  \footnotesize 0.484&  \footnotesize 0.037\\
\hline

\footnotesize Topical Chat & \footnotesize - &  \footnotesize Wiki-727K &  \footnotesize 0.507 &  \footnotesize 0.503 &  \footnotesize 0.023 &  \footnotesize0.516&  \footnotesize 0.510&  \footnotesize 0.031\\
\footnotesize Topical Chat & \footnotesize BOLT & \footnotesize Wiki-727K & \footnotesize 0.511 & \footnotesize 0.514 & \footnotesize 0.199 &  \footnotesize 0.520&  \footnotesize 0.521&  \footnotesize 0.210\\
\hline
\end{tabular}
}

\vspace{-0.5em}
\caption{Comparing the F1 scores with different pre-training \& fine-tuning scenarios using re-weighting and Focal Loss, we conclude that Focal loss stands out to be a much a robust alternative than re-weighted cross-entropy loss for topic segmentation of unstructured texts.The best F1 scores for \textcolor{cyan}{\bf{BOLT}}, \textcolor{purple}{\bf{Topical Chat}}, and \textcolor{orange}{\bf{Wiki-727K}}, per model are highlighted.}
\label{tab:exp10}

\end{table*}
\end{center}

\end{document}